\documentclass[runningheads]{llncs}
\usepackage[T1]{fontenc}
\usepackage{graphicx}
\usepackage{multirow}
\usepackage{amsmath}
\usepackage{algpseudocode}
\usepackage{algorithmicx,algorithm}

\usepackage{diagbox}
\usepackage[table,xcdraw]{xcolor}
\usepackage[font=small,labelfont=bf]{caption}
\usepackage[noadjust]{cite}

\usepackage{easyReview} 
\begin{document}
%
    \title{FFAD: A Novel Metric for Assessing Generated Time Series Data Utilizing Fourier Transform and Auto-encoder}

%
\titlerunning{FFAD: A Novel Metric for Assessing Generated Time Series Data}
%
\author{Yang Chen\thanks{Corresponding author: Yang Chen \\     ychen113@student.gsu.edu} \and Dustin J. Kempton \and Rafal A. Angryk }

\authorrunning{Yang Chen et al.}

\institute{ Georgia State University, Atlanta, GA 30302, USA \\ \email{\{ychen113\}@student.gsu.edu, \{dkempton1, angryk\}@cs.gsu.edu}}
%
%
%
\maketitle 
%
\begin{abstract}
The success of deep learning-based generative models in producing realistic images, videos, and audios has led to a crucial consideration: how to effectively assess the quality of synthetic samples. While the Fr\'{e}chet Inception Distance (FID) serves as the standard metric for evaluating generative models in image synthesis, a comparable metric for time series data is notably absent. This gap in assessment capabilities stems from the absence of a widely accepted feature vector extractor pre-trained on benchmark time series datasets. In addressing these challenges related to assessing the quality of time series, particularly in the context of Fr\'{e}chet Distance, this work proposes a novel solution leveraging the Fourier transform and Auto-encoder, termed the Fr\'{e}chet Fourier-transform Auto-encoder Distance (FFAD). Through our experimental results, we showcase the potential of FFAD for effectively distinguishing samples from different classes. This novel metric emerges as a fundamental tool for the evaluation of generative time series data, contributing to the ongoing efforts of enhancing assessment methodologies in the realm of deep learning-based generative models.

\end{abstract}
\section{Introduction}

    Deep learning has shown remarkable success in numerous tasks and domains, highlighting its effectiveness in tackling complex challenges. Notably, it particularly excels in generative models such as Generative Adversarial Networks (GANs) \cite{goodfellow2014generative}, Conditional GANs (CGANs) \cite{mirza2014conditional}, and Variational Auto-encoders (VAEs) \cite{Kingma2014}, showcasing their capacity to produce realistic images, artwork, and time series data \cite{Radford2016UnsupervisedRL, zhu2017unpaired, chen2022CGAN}. Automatic evaluation metrics such as Inception Score (IS) and Fr\'{e}chet Inception Distance (FID) have been introduced to assess the quality and diversity of generative samples \cite{Salimans2016IS, heusel2017gans}. The IS utilizes the pre-trained image classification network, such as Inception v3 \cite{inception_v3}, to calculate the conditional distribution and the marginal distribution on synthetic samples. Addressing the limitation of IS in considering only the generated samples, Fr\'{e}chet Inception Distance (FID) has become the preferred measure compared (more details regarding FID are provide in Section~\ref{f-distance}). FID is commonly employed for the evaluation of images \cite{chen2023Examine}, and may not be as directly applicable to sequential data, such as text, audio, or time series, due to the absence of a widely accepted pre-trained model designed for extracting feature vectors from time series data. However, the demand for such a metric for assessing the generated time series is even more necessary, given the challenges in evaluating it through visual inspection compared to image data. In this work, we introduce a novel metric called Fr\'{e}chet Fourier-transform Auto-encoder Distance (FFAD). The metric combines the use of a Fourier Transformation with an Auto-encoder methodology, with the primary objective of assessing the quality of synthetic time series samples. Our solution leads to the following contributions:
        \begin{itemize}
        \item Determining a suitable number of frequency components using Fourier Transform to convert data from time to frequency domain, facilitating standardized representation and mitigating the impact of varying lengths of time series datasets.
        \item Showing the effectiveness of compressed representation achieved by training a general Auto-Encoder on an extensive set of time series datasets.
        \item Establishing the FFAD score as an innovative and effective metric in distinguishing data across various classes, establishing it as a fundamental tool for the evaluation of generative time series data.
        \end{itemize}



\section{Related Work}

    \subsection{Fourier Transform}

        The Fourier transform, an integral mathematical technique applied extensively in signal processing, mathematics, and diverse scientific disciplines, is employed to analyze and depict functions or signals within the frequency domain. Its primary objective is the dissection of complex signals into constituent sinusoidal components. This process involves the decomposition of time-domain signals into their corresponding frequency components, providing a comprehensive understanding of the varied frequency contributions comprising the signal. 

        Numerous Fourier Transform implementations have been proposed, including the Discrete Fourier Transform (DFT), Fast Fourier Transform (FFT), Short-Time Fourier Transform (STFT), and Windowed Fourier Transform (WFT). The DFT is widely employed in digital signal processing when dealing with discrete, sampled data. The FFT has an improved computation time compared to the straightforward evaluation of the DFT, making it widely used in applications such as signal processing, image analysis, and many scientific computations \cite{cooley1969fft}. The STFT is used for analyzing how the frequency content of a signal changes over time. It involves applying the Fourier Transform to short, overlapping segments of a signal to capture time-varying characteristics \cite{benesty2011speech}. Similar to the STFT, the WFT involves applying the Fourier Transform to localized sections of a signal using window functions. This allows for better frequency localization \cite{kemao2004stft}. Our primary focus in this work is on utilizing FFT, given its higher computational efficiency compared to other methods.

    \subsection{Auto-encoder}
        Auto-encoders, a class of neural networks, have gained significant attention in the domain of deep learning and unsupervised learning. A typical auto-encoder consists of an encoder and a decoder. The encoder compresses input data into a lower-dimensional representation, while the decoder reconstructs the original input from this code. During training, the network learns to minimize the difference between the input and the reconstructed output, facilitating the extraction of meaningful features in the encoded representation. 
        
        The wide-ranging applications and adaptability of auto-encoders have made them a crucial component of modern deep learning and data representation techniques, with a multitude of innovative approaches and architectures continually emerging from the research community. In addressing specific data types, Convolutional Auto-encoders are customized for image data through the integration of convolutional layers \cite{mao2016image}. Similarly, Recurrent Auto-encoders are devised for sequential data, such as time series, employing recurrent units \cite{cho2014learning}. In this work, we utilize the Recurrent Auto-encoder, considering frequency components as sequential data. 
    
    \subsection{Fr\'{e}chet Distance} \label{f-distance}
        The Fr\'{e}chet distance provides a way for measuring the similarity between curves \cite{2014frechet}. Introduced in 2017, the Fr\'{e}chet Inception Distance (FID) score is the current standard metric for evaluating the quality of generative models in image generation. Using the feature vectors derived from the Inception v3 model \cite{inception_v3}, FID calculates the distance between real and generated images. Specifically, the final pooling layer preceding the classification of output images is used to capture computer-vision-specific features of an input image. In practice, each input image is represented as a feature vector. $X$ and $Y$ are feature vectors of the real and synthetic samples. Then, multivariate FID can be computed based on the formulation in Eq.~\ref{eq:mFID} \cite{DOWSON1982450}. $\mu_{X}$ and $\mu_{Y}$ are the vector magnitudes $X$ and $Y$, respectively. ${Tr(.)}$ is the trace of the matrix, while $\Sigma_{X}$ and $\Sigma_{Y}$ are the covariance matrices of $X$ and $Y$. Lower FID values indicate higher quality and diversity in synthetic samples.
        
        \begin{equation}
        \text{Score} = ||\mu_{X} - \mu_{Y}||^{2} + \text{Tr}(\Sigma_{X}+\Sigma_{Y}-2\sqrt{\Sigma_{X}\Sigma_{Y}})
        \label{eq:mFID}
        \end{equation}

        Moving beyond image generating models, the authors of \cite{preuer2018frechet} introduced the Fr\'{e}chet ChemNet Distance (FCD) as a evaluation metric for generative models in the context of molecular structures relevant to drug discovery. Diverging from the FID, FCD derives feature vector representations for each molecules by utilizing the penultimate layer of the ChemNet \cite{goli2020chemnet}. Subsequently, the Fr\'{e}chet Distance is computed based on the distribution of real and generative samples.

\section{Methodology} \label{sec:methodo}

    Our primary objective is to train an Auto-encoder by utilizing a variety of time series datasets. Subsequently, we intend to employ the Encoder component to generate a lower-dimension representation for any given time series, whether it is a real-world dataset or a synthetic one produced by a generative model, such as a conditional GAN. Additionally, we will calculate the FFAD score to measure the dissimilarity between the distributions of a pair of time series datasets, whether they belong to different categories or involve a comparison of real and synthetic samples.

\begin{figure}[ht!]
        \centering
        \includegraphics[scale=0.50]{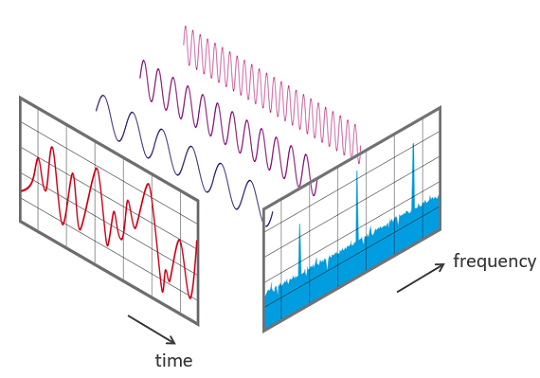}
        \caption{An example of Viewing a time signal in both the time and frequency domains utilizing Fourier Transform. \protect\footnotemark} 
        \label{fig:fft}
    \end{figure}

    To effectively train an Auto-encoder for time series data, the foremost challenge to address is handling datasets with varying sequence lengths. Within the UCR dataset collection, the time series datasets can range in length from 15 to 2844 data points, with an average length of 537 time steps. To tackle this significant variability, we utilize Fourier Transformation as a preprocessing step for the original UCR time series. This choice is guided by two primary reasons: (1) Ensuring all time series data with a consistent input length for training RNN-based auto-encoder. Maintaining a uniform input length eliminates the need for padding the variable-length inputs, resulting in increased time efficiency. (2) Shifting from time domain to frequency domain with Fourier Transform while preserving essential features, as shown in Fig.~\ref{fig:fft}. This transformation enables us to represent the original time series by selecting a suitable number of sine components, which are not only appropriate but also fewer in number compared to the original sequence length.

    \footnotetext{Image source: www.nti-audio.com/en/support/know-how/fast-fourier-transform-fft}

\begin{figure}[hbt!]
        \centering
        \includegraphics[scale=0.3]{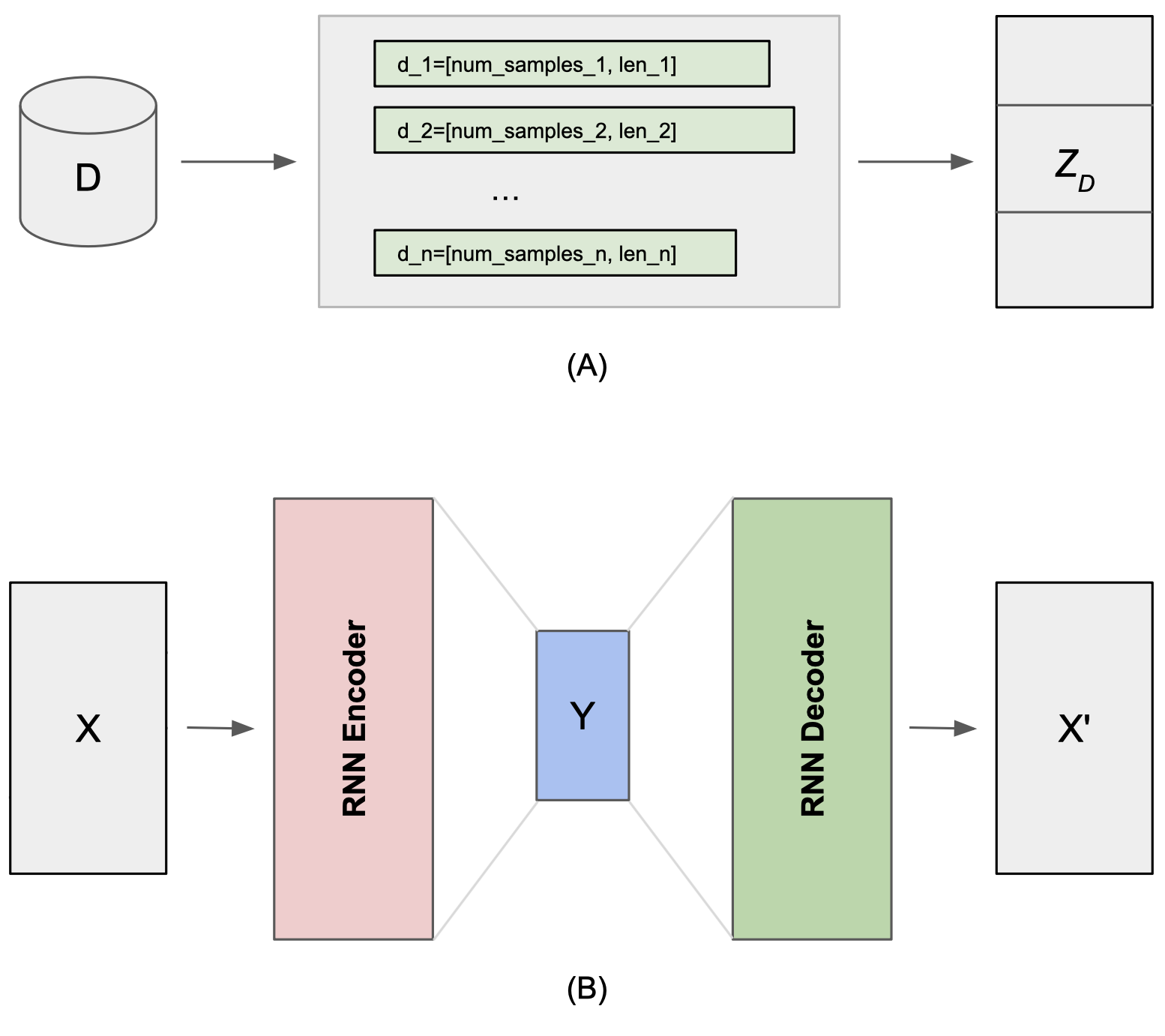}
        \caption{Sub-figure (A) illustrates the procedure of employing Fourier Transformation as a preprocessing step for the original time series data, ensuring a consistent length for all datasets. Sub-figure (B) outlines the training procedure of the autoencoder.} 
        \label{fig:methodol_arch}
    \end{figure}

    Consider a collection of time series datasets denoted as $D=\{d_{1}, d_{2}, ..., d_{n}\}$, as shown in (A) of Fig.~\ref{fig:methodol_arch}. Each individual dataset $d_{i}$ has its own length $len_{i}$ and number of samples $|d_{i}|$. As a result, each $d_{i}$ can be represented as a matrix with dimensions $[|d_{i}|, len_{i}]$. Through the utilization of the Fourier transformation, we can convert each dataset from the time domain to the frequency domain. Assuming a selection of $m$ frequency components, each datasets yields a Fourier-transform (FT) representation matrix ($Z_{d_{i}}$) with the shape of $[|d_{i}|, m]$. Choosing the same $m$ value for all datasets enables the concatenation of the FT representation matrices into a larger matrix ($Z_{D}$) with the shape of $[N, m]$ (i.e., $N=\sum^{n}_{i=1}|d_{i}|$), as shown in Eq.~\ref{eq:ZD}. Within $Z_{D}$, each element (e.g., $z^{k}_{i, j}$) represents $k$ as the dataset index, $i$ as the sample index within each dataset, $|d{i}|$ indicating the total number of samples for a specific dataset, and $j$ as the index of frequency components. 

    \begin{equation}
        Z_{D} = [Z_{d_{1}}, Z_{d_{2}}, ..., Z_{d_{n}}] = 
        \begin{bmatrix}
            z^{1}_{1, 1}       & z^{1}_{1, 2}       & ... & z^{1}_{1, m}       \\
            z^{1}_{2, 1}       & z^{1}_{1, 2}       & ... & z^{1}_{1, m}       \\
            ...                & ...                & ... & ...                \\
            z^{1}_{|d_{1}|, 1} & z^{1}_{|d_{1}|, 2} & ... & z^{1}_{|d_{1}|, m} \\
            z^{2}_{1, 1}       & z^{2}_{1, 2}       & ... & z^{2}_{1, m}       \\
            ...                & ...                & ... & ...                \\
            z^{n}_{1, 1}       & z^{N}_{1, 2}       & ... & z^{N}_{1, m}       \\
            ...                & ...                & ... & ...                \\
            z^{n}_{|d_{n}|, 1} & z^{n}_{|d_{n}|, 2} & ... & z^{n}_{|d_{n}|, m} \\
        \end{bmatrix}
        \xrightarrow{shape}{[N, m]}
    \label{eq:ZD}
    \end{equation}

    An aspect that requires investigation pertains to the nature of Fourier transform results, which manifest as complex numbers, encapsulating both magnitude and phase information. Since we employ Keras, which primarily supports real-valued computations for neural networks, and doesn't have native support for complex numbers in its core operations, so the pre-preocessing for complex numbers is needed. Considering each row in $Z_{D}$ as list of frequency components $z=[z_{1}, z_{2}, ..., z_{m}]$ (where $z_{i}=a+bj$), we separate the real ($a$) and imaginary parts ($bj$) by organizing them into a two-dimension array $z^{'}=[[a_{1}, b_{1}], [a_{2}, b_{2}], ..., [a_{m}, b_{m}]]$, characterized by a shape of $[m, 2]$. The ultimate shape of the matrix $Z_{D}$ will be $[N, m, 2]$.

    In the implementation of the Recurrent Auto-encoder, we employ the Gated Recurrent Unit (GRU) for both the Encoder and Decoder components. GRU is selected for its enhanced performance in processing long sequences, by minimizing the risk of the gradient vanishing problem \cite{chung2014empirical}. We adopt a mini-batch approach to train the Auto-encoder effectively. During each iteration, a batch comprising a batch-size of samples (referred to as $X$) from the matrix $Z_{D}$ serves as input for the Encoder component. The output of the Encoder, denoted as $Y = \text{Encoder}(X)$, represents a significantly compressed representation compared to $X$. For the Encoder training, we require another integral component known as the Decoder. This Decoder takes $\text{Encoder}(X)$ as its input and generates an output represented by $X' = \text{Decoder}(Y) = \text{Decoder}(\text{Encoder}(X))$. The primary goal of the Decoder is to reconstruct the initial input data $X$. Therefore, the overarching objective function of the entire model is to minimize the error between $X'$ and $X$. In practice, we utilize the Mean Square Error (MSE) as the training criterion, as shown in Eq.~\ref{eq:loss_func}.

    \begin{equation}
    \begin{aligned}
        &\text{Loss} = \left(\frac{1}{|batch|} \sum_{i=1}^{|batch|} (X'_i - X_i)^{2}\right)\xrightarrow{}\text{Minimized}, \\
        & \quad where \quad X'_i = \text{Decoder}(Y_i) = \text{Decoder}\left(\text{Encoder}(X_i)\right)
    \end{aligned}
    \label{eq:loss_func}
    \end{equation}

    After completing the training of the auto-encoder, the focus shifts to retaining solely the Encoder component. Taking a binary dataset $d_{i}$ as a case study, we have two sample sets: $S_{\text{pos}}$ representing the positive class and $S_{\text{neg}}$ signifying the negative class. Employing the Encoder, which has been effectively trained, we generate encoded representations denoted as $Y_{\text{pos}}$ and $Y_{\text{neg}}$ for the positive and negative sets, respectively. Then, the FFAD score can be calculated to measure the similarity between $S_{\text{pos}}$ and $S_{\text{neg}}$, as illustrated in Eq.~\ref{eq:FID-like-score}, 

    \begin{equation}
    \text{FFAD Score} = ||\mu_{Y_{\text{pos}}} - \mu_{Y_{\text{neg}}}||^{2} + \text{Tr}(\Sigma_{Y_{\text{pos}}}+\Sigma_{Y_{\text{neg}}}-2\sqrt{\Sigma_{Y_{\text{pos}}}\Sigma_{Y_{\text{neg}}}})
    \label{eq:FID-like-score}
    \end{equation}

    Here, $Y_{\text{pos}}$ and $Y_{\text{neg}}$ serve as the encoded representations, while $\mu_{Y_{\text{pos}}}$ and $\mu_{Y_{\text{neg}}}$ correspond to the vector magnitudes of $Y_{\text{pos}}$ and $Y_{\text{neg}}$, respectively. The function ${Tr(.)}$ denotes the trace of the matrix, with $\Sigma_{Y_{\text{pos}}}$ and $\Sigma_{Y_{\text{neg}}}$ representing the covariance matrices of $Y_{\text{pos}}$ and $Y_{\text{neg}}$. Lower values within this equation indicate a higher similarity between the two input sets.

    \begin{algorithm}[h!]
	\caption{Inverse Fourier Transform}
	\label{Algo:Inverse-FT}
	\begin{flushleft}
	\textbf{Input:} frequencies  // a list of freq components \\ 
	                \hspace*{\algorithmicindent} \hspace*{\algorithmicindent} 
                    length // the original time series' length \\
    \textbf{Output:} rec\_ts // The reconstructed time series.\\
    \end{flushleft}
    \begin{algorithmic}[1]
        \State Set rec\_ts to an array of zeros with length elements
        \State Set index to an array containing values of [0, length) with increments of 1.0
        \For {$i = 1$ to $length$}
            \State Set rec\_ts to 0
            \State Set index[i] to i * (2 * pi) / length 
        \EndFor

        \For {$k$, $p$ in enumerate($frequencies$)}
            \State If k is not equal to 0, then multiply p by 2
            \State Separate the real and imaginary components of p as a+bj
            \For {$j=1$ to $length$}
                \State Add a * cos(k * index[j]) to rec\_ts[j]
                \State Subtract b * sin(k * index[j]) from rec\_ts[j]
            \EndFor    
        \EndFor
	\end{algorithmic} 
\end{algorithm}

    Furthermore, we implement the Inverse Fourier Transform to verify the reconstruction capability of the transformed data. As shown in \ref{Algo:Inverse-FT}, the Inverse Fourier Transform function requires two parameters: $frequencies$ and $length$. The $frequencies$ parameter is anticipated to be a list or array containing the Fourier transform coefficients of the sequence. On the other hand, $length$ signifies the length of the original time series. The goal of the Inverse Fourier Transform is to merge these Fourier coefficients to reconstruct the original time series.

\section{Experiments and Results}
    
    \subsection{Dataset} \label{sec:dataset}
        Our experimentation focused primarily on two widely recognized public datasets: UCR and SWAN-SF. The UCR Time Series Classification/Clustering Repository, curated by the University of California, Riverside (UCR), stands as a prominent resource in the fields of time series analysis and data mining. It plays a pivotal role as a benchmark for the development and assessment of algorithms and models tailored for time series classification, clustering, and related tasks. The UCR datasets consist of 117 datasets of fixed length. In our study, we specifically utilized 97 out of the 117 datasets, as our Fourier transform-based methodology was found to be incompatible with the remaining 20 datasets. 

        SWAN-SF \cite{angryk2020multivariate} refers to a comprehensive, multivariate time series dataset extracted from solar photospheric vector magnetograms in HMI Active Region Patch (HARP) data, publicly available as the Space-weather HMI Active Region Patch (SHARP) series \cite{hoeksema2014, bobra2014helioseismic} at the Harvard Dataverse Repository \cite{angryk2020multivariate-data}. The benchmark dataset is made up of five temporally non-overlapping partitions spanning the period from May 2010 through August 2018. Each multivariate time series is labeled based on the strongest flare event observed in the 24-hour prediction window. The SWAN-SF has 51 field parameters, but many of the parameters are highly correlated, and therefore, this work will focus on four representative parameters, including TOTUSJH, ABSNJZH, SAVNCPP, and TOTBSQ (for the definition of parameters see Table 1 in \cite{angryk2020multivariate}). 
    

    \subsection{Experiments and Analysis}
        \subsubsection{A. Transforming Data with Fourier Transform} 
            \quad \\
            
            We conducted an experiment to evaluate the capacity of various frequency components in representing the original sequences through Fourier transformation. Our investigation involved assessing the reconstruction capability across different numbers of frequency components: $\{1, 2, 3, 5, 10, 15, 20, 30\}$. To conduct this analysis, we selected the SWAN-SF dataset, focusing on four parameters from partition-1. This selection was deliberate for two key reasons: (1) SWAN-SF represents a real-world dataset ideal for reconstruction studies, and (2) its extensive collection of time series encapsulates the inherent complexity often observed in such data.

\begin{figure}[hbt!]
                \centering
                \includegraphics[scale=0.3]{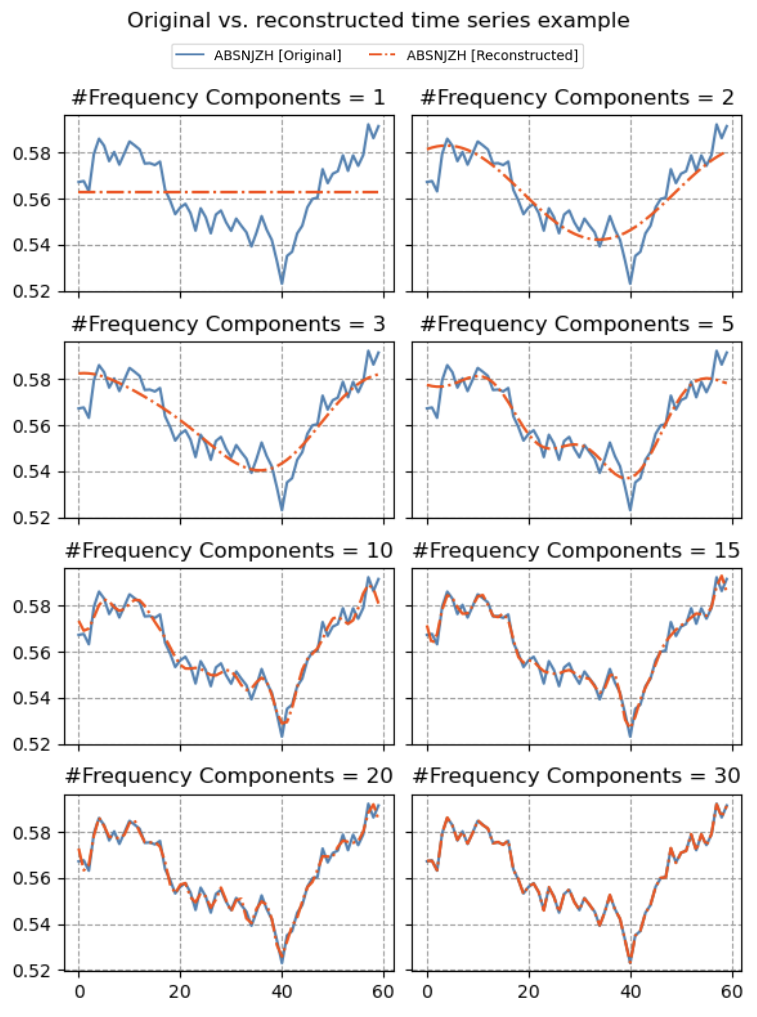}
                \caption{Shows the original time series and reconstructed time series utilizing different number of frequency components. This example is sourced from partition-1 of SWAN-SF.}
                \label{fig:Exp-A-example}
            \end{figure}

            To assess the reconstruction capability of the transformed data, we employed the Inverse Fourier Transform (Inverse-FT) as described in Section \ref{sec:methodo}. Fig.~\ref{fig:Exp-A-example} illustrates the reconstruction examples for the \text{ABSNJZH} parameter using varying numbers of frequencies, ranging from 1 to 30. When the number of frequency components is between 1 and 5, the Inverse-FT can only approximate the major trend of the input time series in a coarse manner. However, with the utilization of more frequency components (i.e., {10, 15, 20, 30}), the restored time series becomes more refined, and the Inverse-FT can fully recover the input time series using all 30 components (considering the input time series comprises 60 time steps). 
            
\begin{figure}[hbt!]
                \centering
                \includegraphics[scale=0.5]{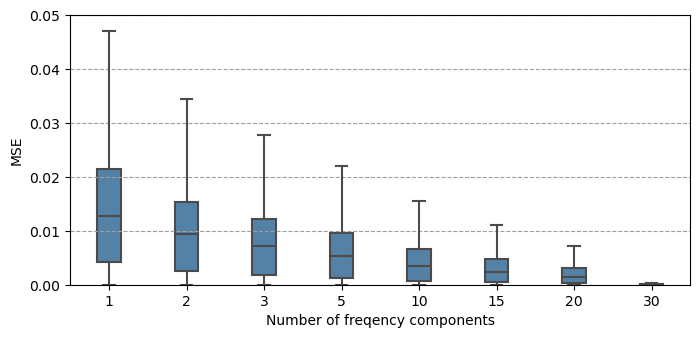}
                \caption{The results provide a comprehensive evaluation of the reconstruction performance on SWAN\_SF partition-1.} 
                \label{fig:Exp-A-result}
            \end{figure}

            To conduct a comprehensive evaluation of the reconstruction performance, we employ Mean Square Error (MSE) as the evaluation metric, as discussed in Section~\ref{sec:methodo}. The corresponding results are depicted in Fig.~\ref{fig:Exp-A-result}. It is evident from the figure that the Mean Squared Error (MSE) decreases as more frequency components are incorporated into the reconstruction process. 
            We conclude that employing 20 components achieves a favorable equilibrium between compressed representation and reconstruction capability for subsequent experiments.

        \subsubsection{B. Training Auto-encoder and Model Selection Criteria}
            \quad \\
            
            To train a unified Auto-encoder, we combined 97 UCR datasets with the SWAN-SF dataset. The training dataset comprises the training sets from the 97 UCR datasets, as well as partition-1 of SWAN-SF, which includes 4 parameters. Similarly, the testing dataset includes the testing sets from the 97 UCR datasets, along with partition-2 of SWAN-SF, containing 4 parameters. We tune the performance of Auto-encoder model by setting different hyper-parameters, i.e. GRU hidden sizes: $\{5, 10, 20, 30\}$, learning rates: $\{0.1, 0.01, 0.001, 0.0001\}$, batch sizes: $\{256, 512, 1024\}$. Empirically, we concluded our optimal hyper-parameter setting with the GRU hidden size of $20$, the learning rate of $0.001$, the batch size of $512$. The model was trained with $5000$ epochs.

            \begin{figure}[hbt!]
                \centering
                \includegraphics[scale=0.6]{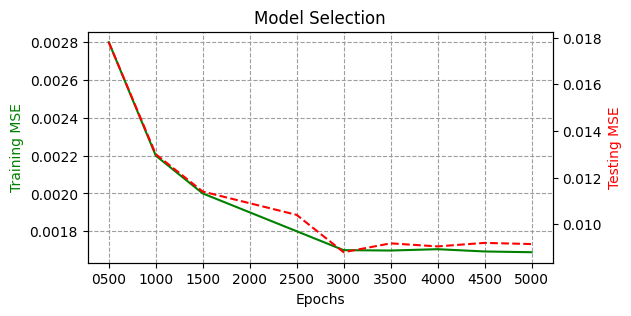}
                \caption{Shows the procedure of selecting the Auto-encoder model by calculating Mean Square Error (MSE) as the evaluation metric every 500 epochs, and identifies that the optimal model is achieved at the 3000th epoch.}
                \label{fig:Exp-B-result}
            \end{figure}

            \begin{figure}[hbt!]
                \centering
                \includegraphics[scale=0.18]{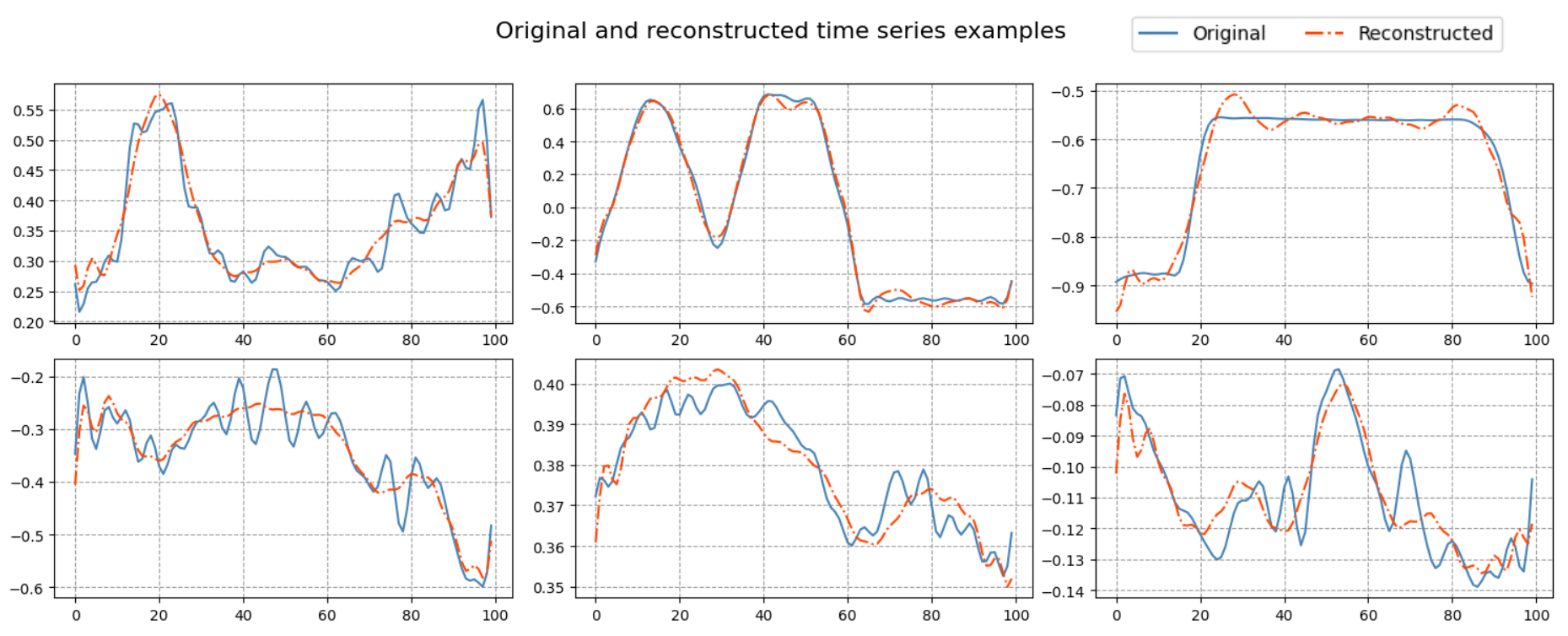}
                \caption{Displays six pairs of original and reconstructed time series. The examples are selected from the UCR and SWAN-SF.} 
                \label{fig:Exp-B-example}
            \end{figure}

            To perform model selection during the Auto-encoder's training, we utilize Mean Square Error (MSE) as the evaluation metric over every 500 epochs. More specifically, we randomly select 10,000 from both the training and testing sets to compute the MSE, and the outcomes are depicted in Fig.~\ref{fig:Exp-B-result}. Analyzing the training and testing curves, we identify that the optimal model is achieved at the 3000-th epoch. Additionally, Fig.~\ref{fig:Exp-B-example} shows six pairs of original and reconstructed time series examples, selected from the testing test.

        \subsubsection{C. Assessing the Effectiveness of the FFAD Score}
            \quad \\

            After the completion of Auto-encoder training, the Encoder component can be effectively utilized in FFAD-based evaluation. Consider an arbitrary binary dataset, segmented into a training set and a testing set, each comprising two classes (e.g., class-0 and class-1), resulting in a total of four sub-datasets. The FFAD score is then computed using these four sub-datasets, pairing any two of them to yield six scores. Notably, two out of the six scores are determined based on the same classes (e.g., train-0 vs. test-0 and train-1 vs. test-1). The remaining four scores are determined from datasets representing different classes (e.g., train-0 vs. train-1, train-0 vs. test-1, train-1 vs. test-0, and test-0 vs. test-1). Table.~\ref{table:Experiment-C-results} summarizes the FFAD results for ten binary UCR datasets. Upon observing the results, it is evident that FFAD scores for same-class comparisons are notably lower than those for different-class comparisons. This observation implies the effectiveness of FFAD in distinguishing between samples from the same or different classes. In addition, this capability will be particularly valuable when evaluating generative models for binary- or multi-class time series generation, such as conditional GAN. This feature not only enables the assessment of the realism of generated samples but also facilitates checking if the generated samples correspond to their class information.

            \begin{table}[hbt!]
\caption{The table reported the FFAD scores for ten binary UCR datasets, with each score representing the average from five repeated experiments. The actual scores are presented in scientific notation, multiplied by $10^{-5}$.}
\label{table:Experiment-C-results}
\begin{center}
\begin{tabular}{    |p{2.4cm}<{\raggedright}
                    |p{1.5cm}<{\centering}
                    |p{1.5cm}<{\centering}
                    |p{1.5cm}<{\centering}
                    |p{1.5cm}<{\centering}
                    |p{1.5cm}<{\centering}
                    |p{1.5cm}<{\centering}|}
\hline
\multirow{2}{*}{\begin{tabular}[c]{@{}l@{}} Binary\\ Datasets\end{tabular}}
& \multicolumn{2}{c|}{Same-class} & \multicolumn{4}{c|}{Different-class}
\\ \cline{2-7} 
& \multicolumn{1}{c|}{\begin{tabular}[c]{@{}c@{}}train-0 vs.\\ test-0\end{tabular}} 
& \multicolumn{1}{c|}{\begin{tabular}[c]{@{}c@{}}train-1 vs.\\ test-1\end{tabular}} 
& \multicolumn{1}{c|}{\begin{tabular}[c]{@{}c@{}}train-0 vs.\\ train-1\end{tabular}} 
& \multicolumn{1}{c|}{\begin{tabular}[c]{@{}c@{}}train-0 vs.\\ test-1\end{tabular}} 
& \multicolumn{1}{c|}{\begin{tabular}[c]{@{}c@{}}train-1 vs.\\ test-0\end{tabular}} 
& \multicolumn{1}{c|}{\begin{tabular}[c]{@{}c@{}}test-0 vs.\\ test-1\end{tabular}} 
\\ \hline
1. BeetleFly & 67.81 & 64.07 & 174.01 & 146.74 & 335.66 & 243.5 \\ \hline
2. BirdChicken & 40.34 & 18.11 & 43.9 & 52.76 & 64.43 & 71.39 \\ \hline
3. ECG200 & 202.86 & 28.7 & 1431.86 & 1336.32 & 1298.21 & 1090.2 \\ \hline
4. ECGFiveDays & 0.75 & 5.09 & 24.45 & 15.08 & 20.84 & 11.1 \\ \hline
5. GunPoint & 22.75 & 15.66 & 426.84 & 314.97 & 327.61 & 219.7 \\ \hline
6. Lightning2 & 1.44 & 1.1 & 4.47 & 3.71 & 4.47 & 1.86 \\ \hline
7. Strawberry & 1.37 & 10.74 & 81.88 & 48.43 & 102.79 & 62.11 \\ \hline
8. TwoLeadECG & 10.93 & 4.52 & 47.0 & 29.38 & 34.54 & 18.59 \\ \hline
9. Wafer & 2.52 & 1.17 & 11.21 & 11.16 & 14.77 & 11.32 \\ \hline
10. Yoga & 3.3 & 4.15 & 21.26 & 13.13 & 15.89 & 12.8 \\ \hline
\end{tabular}
\end{center}
\end{table}

\section{Conclusion}

    In this study, we have introduced a novel metric named the Fréchet Fourier-transform Auto-encoder Distance (FFAD), by seamlessly integrating the Fourier transform and Auto-encoder. Our experimental results demonstrate the effectiveness of FFAD in distinguishing samples from various classes, positioning it as a fundamental tool for evaluating the quality of generative time series data. In our future work, we will focus on utilizing FFAD as the evaluation metric for assessing generative models in the context of flare forecasting tasks within time series generation. We firmly believe that adopting such a comprehensive metric not only addresses immediate challenges related to quality and diversity but also provides avenues for refining existing methodologies in time series generation.

\section*{Acknowledgment}
    It will be provided at publication.

%
%
%
\bibliographystyle{IEEEtran}
\bibliography{bibliography}

%




\end{document}